\documentclass{article}

\PassOptionsToPackage{numbers, compress}{natbib}
    \usepackage[final]{neurips_2022}

\usepackage[utf8]{inputenc} % allow utf-8 input
\usepackage[T1]{fontenc}    % use 8-bit T1 fonts

\usepackage{natbib}

\PassOptionsToPackage{hyphens}{url}
\usepackage{hyperref}
\usepackage{booktabs}       % professional-quality tables
\usepackage{amsfonts}       % blackboard math symbols
\usepackage{nicefrac}       % compact symbols for 1/2, etc.
\usepackage{microtype}      % microtypography
\usepackage{xcolor}         % colors
\usepackage{graphicx}
\usepackage{float}
\usepackage[autostyle]{csquotes}  
\usepackage{subcaption}

\makeatletter
\newcommand\footnoteref[1]{\protected@xdef\@thefnmark{\ref{#1}}\@footnotemark}
\makeatother

\title{Can There be Art Without an Artist?}

\author{%
  Avijit Ghosh \\
  Northeastern University\\
  \texttt{ghosh.a@northeastern.edu} \\
  \And
  Genoveva Fossas \\
  Northeastern University\\
  \texttt{fossas.g@northeastern.edu} \\
}

\begin{document}

\maketitle

\begin{abstract}
  Generative AI based art has proliferated in the past year, with increasingly impressive use cases from generating fake human faces to the creation of systems that can generate thousands of artistic images from text prompts - some of these images have even been ``good'' enough to win accolades from qualified judges. In this paper, we explore how Generative Models have impacted artistry, not only from a qualitative point of view, but also from an angle of exploitation of artists -- both via plagiarism, where models are trained on their artwork without permission, and via profit shifting, where profits in the art market have shifted from art creators to model owners. However, we posit that if deployed responsibly, AI generative models have the possibility of being a positive, new modality in art that does not displace or harm existing artists.
\end{abstract}

% From whiteboard
\section{Introduction}

\begin{figure}[h]
\centering
\includegraphics[width=0.7\textwidth]{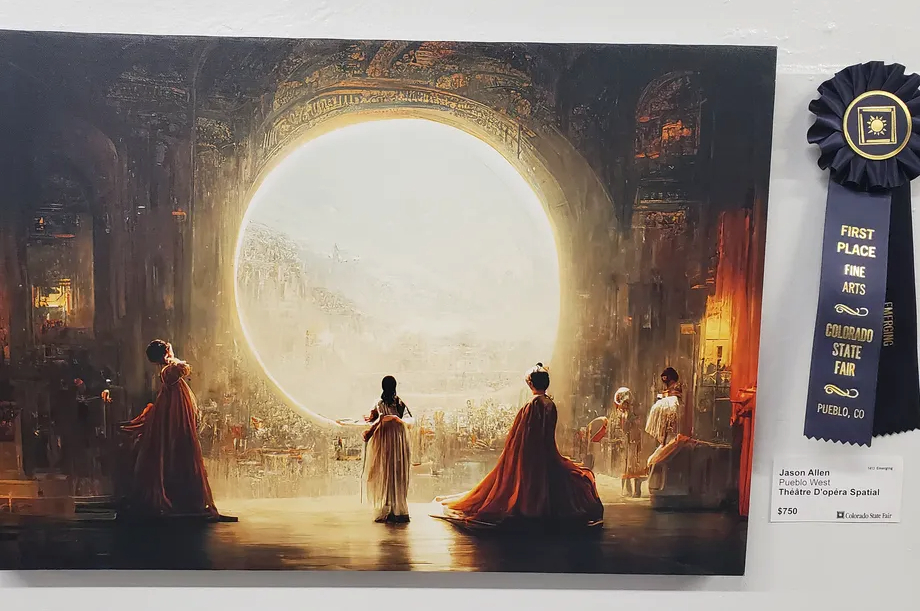}
\caption{Théâtre D’opéra Spatial by Jason Allen}
\end{figure}

Jason Allen, a Colorado based video game designer, caused a stir in the art community when he won first prize in the digital art category at the Colorado State Fair by revealing that his winning piece, \textit{Théâtre D’opéra Spatial}, was created using Midjourney\footnote{\label{note1}\url{https://www.midjourney.com/home/}}, an Generative AI Art model that outputs images given text prompts as inputs. The resulting debate in the art community highlighted the tensions between technological progress in the digital art space, where much like anything else in today's world, machine learning is quickly changing the landscape. This rapid change is felt in the form of economic impacts by digital artists as power and profits shift from artists to model and computational infrastructure owners. Even though Allen admittedly was responsible for inputting the text prompts that ultimately generated the winning piece, not only is that a severely limited scope of artistic effort compared to the other artists that competed with him in the category, but also it is not hard to imagine a future where even the text prompts could be generated by language models, thereby completely dehumanizing the creative artistic process and severely distorting the human perception of the meaning behind an image. We are thus prompted to ask: \textit{can there be art without an artist?}

%% TODO: Fix this paragraph for new sections
In this paper, we look at several emerging tensions in the arena of digital art. In Section \ref{sotagan} we discuss the state of the art on Generative AI Art models, including the latest models and artists using these tools to augment their work. In Section \ref{attrconsent} we highlight how AI models are trained on stolen artwork from the internet without permission, and the lack of consensus about legal recourses available to artists under current copyright regimes. In Section \ref{profitshift} we discuss the emergent profit shifting crisis - profits are shifted from freelance artists to large corporations who own the models, and additionally, artists are also being actively mimicked and displaced from their jobs.  Finally, in Section \ref{aiastool} we discuss how generative AI models can become their own style or modality of art without threatening existing artists, much like photography and digital art did. We end the paper with a call to arms and regulatory suggestions to prevent such a landscape.

\section{The current landscape of Generative AI Art} \label{sotagan}

Generative Art, defined as art created with an autonomous system, has been around for a long time \cite{boden2009generative}, and predates the mainstream adoption of AI. While there is a lot of rules-based art that does not use machine learning, for the purposes of this paper, we limit our focus to Generative AI Models, such as Generative Adversarial Networks (GANs), and their successors, Text-to-Image AI Models.

In the earliest iterations of GAN art, artists would hand-pick the images in their datasets and carefully hone their algorithm manually to generate images that would become a component of or aid in their creation of art. DeepDream, a project headed by Google engineer Alexander Mordvintsev, was among one of the first discoveries of the artistic potential of AI. While the tool was created to understand neural networks \cite{szegedy2015deepdream}, Google engineers discussed the images' beauty in a Google AI blog entry about the tool. The blog entry notes that artists and programmers alike were interested in how these images were generated, so they made the code used to generate the images publicly available. \cite{deepdream} 

StyleGAN \cite{karras2019style}, released by NVIDIA, was one of the paradigm shifting advances in the area of GANs. StyleGANs were able to transfer the art style of one image onto another, and were also used to generate photo-realistic human faces.\footnote{\url{https://thispersondoesnotexist.com/}} Soon after, OpenAI, the maker of large language model (LLM) GPT, GPT-2 and GPT-3 \cite{brown2020language}, released CLIP \cite{radford2021learning}, which connected previous work on image embeddings with advances in LLMs to present a method to jointly train image and text embeddings and massively improved the robustness of image classifiers trained on non-Imagenet data. OpenAI then released DALL.E \cite{ramesh2021zero}, which they described as: \textit{``a 12-billion parameter version of GPT-3 trained to generate images from text descriptions, using a dataset of text–image pairs.''} Several other teams and companies subsequently announced models, including DALL.E's successor, DALL.E-2, Google's Imagen \cite{saharia2022photorealistic}, Midjourney\footnoteref{note1}, and most recently Stable Diffusion.\footnote{\url{https://github.com/CompVis/stable-diffusion}}

% \subsection{Highlight positive use-cases}
% (if we have the space)

\section{The lack of attribution and consent in model training} \label{attrconsent}

State-of-the-art vision AI models have historically been trained on large datasets scraped from the internet. The most popular datasets in the past decade have been ImageNet \cite{deng2009imagenet}, CelebA \cite{liu2015faceattributes}, and Microsoft COCO \cite{lin2014microsoft}, which have been used to power fundamental models such as AlexNet \cite{krizhevsky2017imagenet}, ResNet \cite{he2016deep}, and StyleGAN \cite{karras2019style}. Each of these widely popular datasets have attribution and consent issues. The images in these datasets were scraped from the internet with no heed to copyrights or image licenses. For instance, in a discussion about copyright issues in MS-COCO on GitHub \cite{cocoimages}, user Micah Elizabeth Scott pointed out the following chilling realization: 

\blockquote{
\footnotesize
There are many different Creative Commons licenses represented by the project, and in fact most of the images seem to be released under terms that are not being upheld by the COCO dataset's distribution terms.

For example, just looking at the unlabeled image dataset from 2017, there are 123403 images with license annotations in the JSON, but only 6614 (about 5\%) of these images are released under the unrestricted or USgov licenses. The other images all require attribution, and some of them additional require share-alike, no derivatives, or non-commercial restrictions. As for the attribution requirement, I don't see how that is served- the image database links to the original address each image was retrieved with from Flickr's CDN, but this does not link back to the image's author or include any of the author's metadata.
}

The DALL.E-2, Imagen, Midjourney, and Stable Diffusion models use a newer, bigger dataset called LAION-5B. Previous work \cite{birhane2021multimodal} points out that LAION not only suffers from problematic stereotypes but also blatant ignoring of copyrights that the original creators had posted along with their work. Artists have rightfully started to take note and raise complaints\footnote{\url{https://twitter.com/tinymediaempire/status/1564737777635311616}} that their work is effectively stolen when model trainers train their commercial products on the artists' original work without permission. The theft is not just plagiarism in terms of artistic style, but often verbatim unlicensed reproduction. Deep learning models have been shown to reproduce training data unmodified \cite{carlini2021extracting}, and this is also the case for generative art models\footnote{\url{https://twitter.com/arvalis/status/1558623546879778816}}.

Unfortunately for artists looking to protect their work, there remains significant uncertainty in terms of the availability of legal recourse under the current copyright landscape. For instance, Creative Commons \cite{ccommons} notes that while in the US, the use of published work to train AI is considered fair use, in the EU, the Directive of the Digital Single Market provides an exception for non-commercial entities to train models, but it allows creators to include language that prohibits model training with their work for commercial purposes. This lack of consensus is detrimental to artists seeking to protect their work, especially when their work is published on the internet that spans across jurisdictions.

%TODO: Things to address
% 4.1: I think it’s totally true that midjourney/dalle/etc are pretty much just money printers for their owners. That said, I would have liked to see some more nuanced discussion of where exactly the tradeoff occurs — what kinds of artists, what kinds of art, etc., get “supplant[ed]” (112). Would the money that goes to dalle+ really have otherwise gone to human creators? For example, I know anecdotally of people who play around with dalle+ just for fun, but if they hadn’t spent the money on those credits, they probably wouldn’t have bought an equivalent amount of art. Or, maybe illustrations for online articles, or talk slides, will be AI-generated (and I’ve seen this happen already), but is that replacing the status quo of human artists? or is it replacing the status quo of picking some stock image? 

% Something to think about here.... marketing expense is not something companies really want to spend money on. Its just some thing to minimize cost. Would artists really have gotten this money anyway? (in the modern age?) Maybe it is just picking another stock image.

\section{Shifting Profit Models} \label{profitshift}

While there has been significant technical progress in terms of more sophisticated models producing higher-quality output, the proliferation of generative models has also led to a seismic shift in how incentives work in the arts, moving power and profits away from individual creators and risk fundamentally shifting the perception of artwork.

% TODO: Maybe we should leave NFT mention in here for bored apes stuff?
\subsection{Commercial Model owners reap the profits, not artists} 

Certainly, an argument can be made that AI text to image models are just another tool in the artists' arsenal, however, reality is much more nuanced. Most of the models discussed earlier in this paper are owned and maintained by private companies, and these entities charge for model usage. It costs \$15 to buy 115 credits on DALL.E (one run of the model consumes one credit), while it costs \$30 a month to use Midjourney without restrictions. Dreamstudio\footnote{\url{https://beta.dreamstudio.ai/membership}}, the supported GUI for Stable Diffusion, charges \pounds10 for 1000 art generations. This essentially means that not only are the artists, whose original works were scraped without permission to train these models, not getting profits from this business model, but under the garb of democratization of art, the middlemen, who in this case are companies like OpenAI or Midjourney, are siphoning off money that would have otherwise gone to original art creators. In plainer terms, these companies are profiting off the artistic endeavors of others, and possibly supplanting those human artisans.

\subsection{Displacement of Artists from freelance jobs}

As Generative AI models produce more believable and visually appealing outputs, some companies are moving away from hiring graphic designers for simple tasks. This is especially visible in Alibaba's usage of Generative AI models in their ad campaign. While Alibaba says that its method of producing these ads still relies on human creativity\footnote{\url{https://www.cnbc.com/2018/07/04/alibabas-ai-makes-thousands-of-ads-a-second-but-wont-replace-humans.html}}, it is undeniable that fewer, less efficient human graphic designers will be hired overall. This lack of variety inherent in a smaller group of graphic designers will skew the outputs of their ad designs as well as put several artists out of work.

\begin{figure}[t] 
\centering
\includegraphics[width=0.8\linewidth]{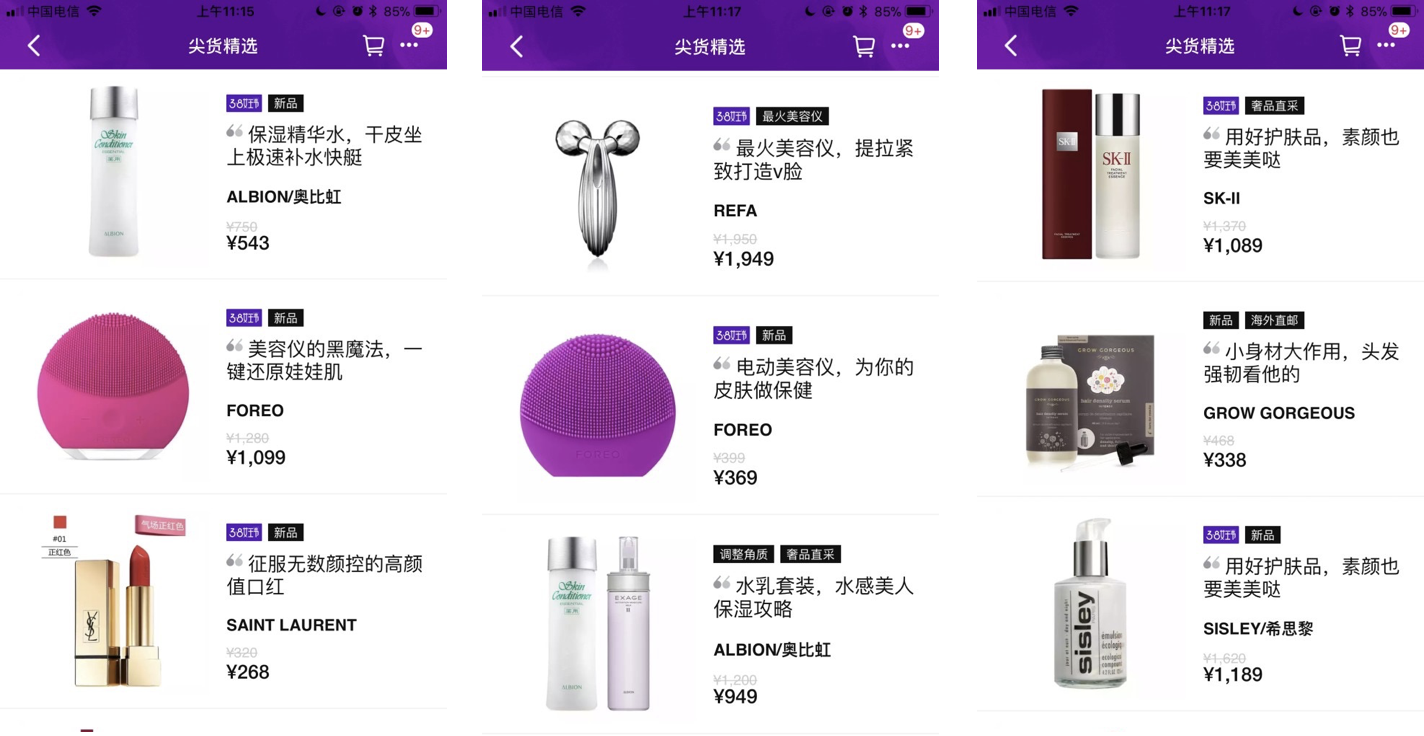}
\caption{AI Generated product ads on Alibaba's website.}
\label{fig:arvalis}
\end{figure}

\subsection{Identity theft via AI Explicitly Mimicking Living Human Artists}

Further development of Generative AI has realized the creation of models specifically meant to closely mimic human artistry. People using Stable Diffusion, for instance, have been able to closely mimic specific artists' styles. One artist called attention to this, and showed a user that was able to use stable diffusion to mimic a large array of artists through with the tool. The artists mimicked are in shown in figure \ref{fig:artists}. The creator of that page has since made it private, but the URL is still able to be seen in relevant spaces on social media\footnote{\url{https://www.notion.so/e2537cbf42c34b7e9a9a4126f81dfd0d}}.

\begin{figure}[h]
    \centering
        \begin{subfigure}{.7\textwidth}\label{fig:artists}.
          \centering
        \raisebox{17mm}{
        \includegraphics[width=\linewidth]{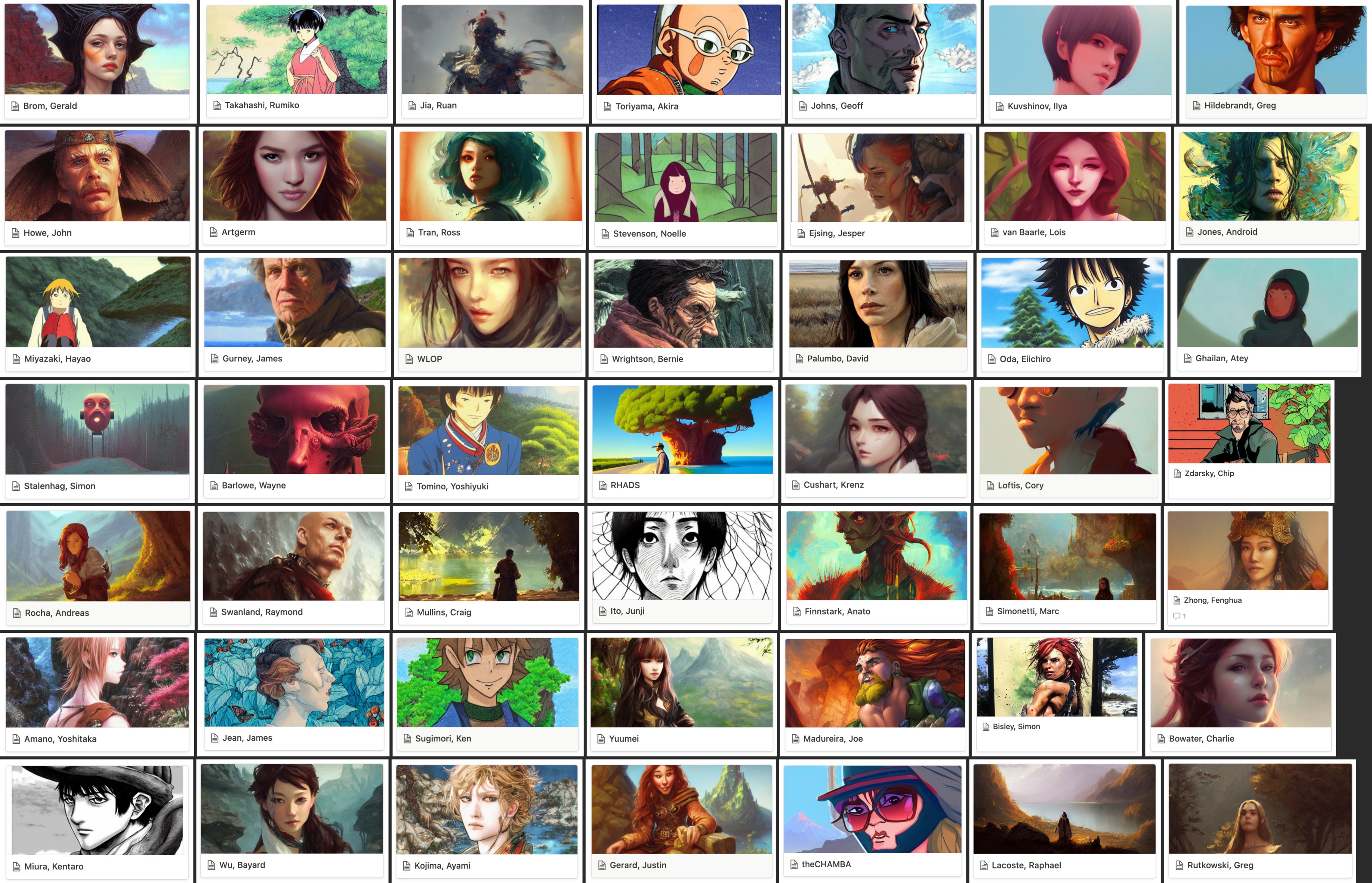}}
        \caption{List of artists a user of stable diffusion was able to mimic.}
        \label{fig:artists}
        \end{subfigure}%
        \hfill 
        \begin{subfigure}{.3\textwidth}
          \centering
        \includegraphics[width=\linewidth]{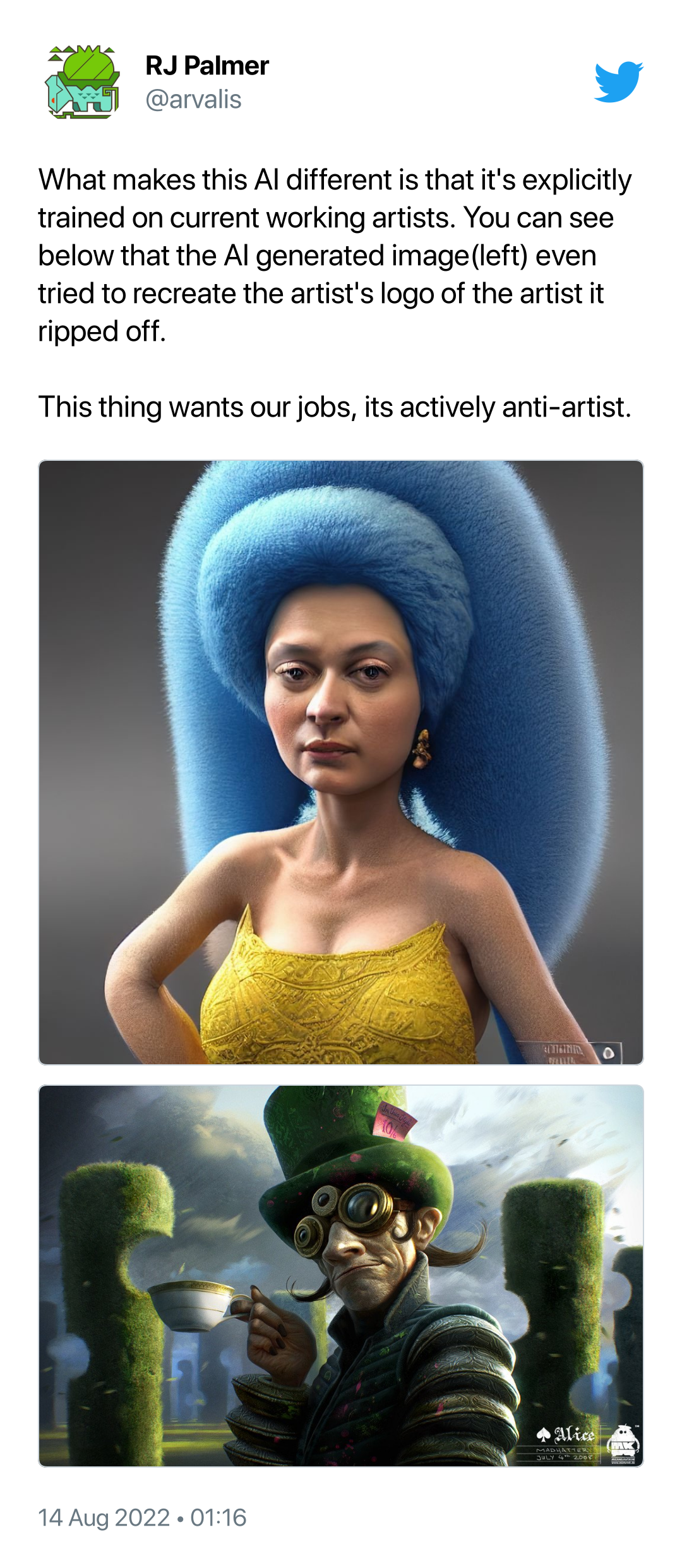}
        \caption{Models sometimes attempt to clone the artist's logo.}
        \end{subfigure}%
        \caption{Models Mimicking Living Human Artists}
    \end{figure}

\section{Artificial Intelligence as an Artistic Tool} \label{aiastool}

When photography was introduced, there were fears that it would completely disrupt and possibly destroy the fine arts market. That did not turn out to be the case, however. Traditional illustration and photography currently coexist in artistic circles and have a unique set of benefits and challenges that result in unique art in both mediums. \cite{moore1992photography} This was also the case for digital art when first introduced; there was a heated debate in the art community about if digital art would cheapen or replace the process of creating traditional art. The introduction of artificial intelligence is fundamentally no different than photography or digital art in this sense.

The growth of digital art as a medium is an especially apt comparison to the introduction of AI. When art-sharing platforms such as DeviantART became popular and digital art tools became more accessible, more people began to create digital art to share publicly with the world \cite{digitalartinterview}. We do not attempt to wade into the debate of what constitutes art itself, as that is completely subjective and a matter of endless debate \cite{tate}. However, in the book \emph{A Philosophy of Computer Art} \cite{philosophyofcomputerart}, there is a description of art that discusses how art has grown and changed with time:

\blockquote{There’s more to art than sitting around and looking pretty. Art works are opportunities for action, and their value depends on the tasks they afford. First comes creative activity, then sometimes performance, and finally appreciation, which includes looking, listening, reading, interacting, interpreting, liking, critiquing, and much else besides.}

Rather than the subjective interpretation of the art being its defining factor, art is instead defined as something that affords creative activity first and foremost. While many artists were genuinely honing their artistry through digital art, some began to trace other artists' work and attempt to profit off of it. Or simply re-posted artists' work claiming it was their own. Direct copying of artists does provide a learning opportunity when handled ethically, but not individual creative expression. In the case of photography and digital art, its the unique capabilities of these mediums that yielded the best works and the most artistic activity in this case. AI art and its generation affords opportunities to create images that no other medium truly could, meaning it is well poised to become its own medium or style. In appendix \ref{app:aiart-positive}, we discuss how several artists use generative models as a part of their process for creative expression as examples of ethical uses of Generative AI to enhance their craft.

\section{Conclusion}

% \vspace{-10mm}

Art as a medium has always molded to fit the society that creates it, from cave drawings created by early humans to fine art painted with quality acrylics, to contemporary technology powered society. That being said, we strongly suggest that the inclusion of AI into art should be performed responsibly in order to avoid the legal, financial, and ethical impacts on human artists. We also believe that AI art is fully capable of becoming a large, unique medium of its own, like photography and digital art did.

Efforts in the area of ethical usage of artists' work have already started to take place in the form of responsible use licensing \cite{contractor2022behavioral}, but the enforcement of usage guardrails should ideally be taken up with a mix of community effort and centralized regulation. We propose the following call to arms to the community: (1) Dataset creators should gain consent before adding art into datasets, à la Deviantart \cite{deviantart}. (2) Model creators should provide tools to unlearn \cite{bourtoule2021machine} training data if the owner of said data revokes consent at a future date. (3) All proposed legislation should support individual artists, not just corporations. (4) Clear, actionable guidelines on transformative use in the light of AI's adoption into art.

% \begin{itemize}
%     \item Dataset creators should gain consent before adding art into datasets, à la Deviantart \cite{deviantart}.
%     \item Model creators should provide tools to unlearn training data if the owner of said data revokes consent at a future date.
%     \item All proposed legislation should support individual artists, not just corporations.
%     \item Clear, actionable guidelines on transformative use in the light of AI's adoption into art.
% \end{itemize}

\section{Acknowledgments}

% \begin{ack}
The authors would like to thank Christo Wilson, Dhanya Lakshmi, Rumman Chowdhury, Neil Turkewitz, and Lucas Do for their constructive feedback and suggestions for improving the manuscript.

% \end{ack}

{\small
\bibliographystyle{plainnat}
\bibliography{ref}
}

\appendix

\section{Appendix}

\subsection{Positive use cases for AI Enabled Art} \label{app:aiart-positive}

The above position piece is largely critical of AI Art, but our position is more nuanced. As long as technology enables artists to do more interesting work without completely displacing them, it is generally a force for good. To that effect, we discuss a few real world examples of artists who have incorporated AI into their own work in a responsible and creative fashion. 

\begin{itemize}
    \item \textbf{Helena Sarin.} Helena creates art by hand, trains GANs on her work to generate new designs, and then brings it back to traditional media like pottery. She therefore uses GAN as a paintbrush to enhance her work. Her work can be found at \url{twitter.com/neuralbricolage}.

    \begin{figure}[H]
    \centering
        \begin{subfigure}{.45\textwidth}
          \centering
          \includegraphics[width=\linewidth]{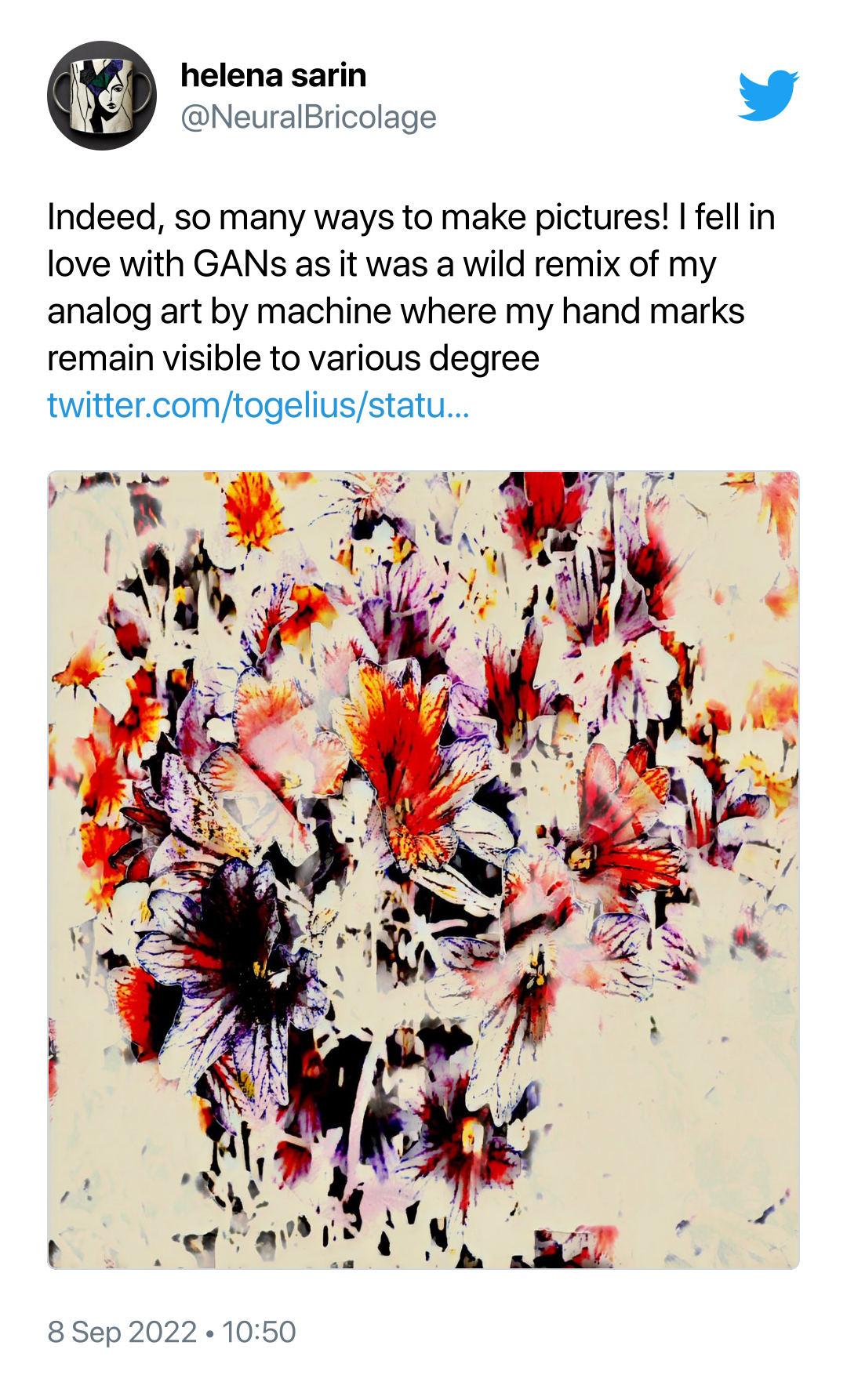}
        \end{subfigure}%
        \begin{subfigure}{.45\textwidth}
          \centering
          \raisebox{6mm}{
          \includegraphics[width=\linewidth]{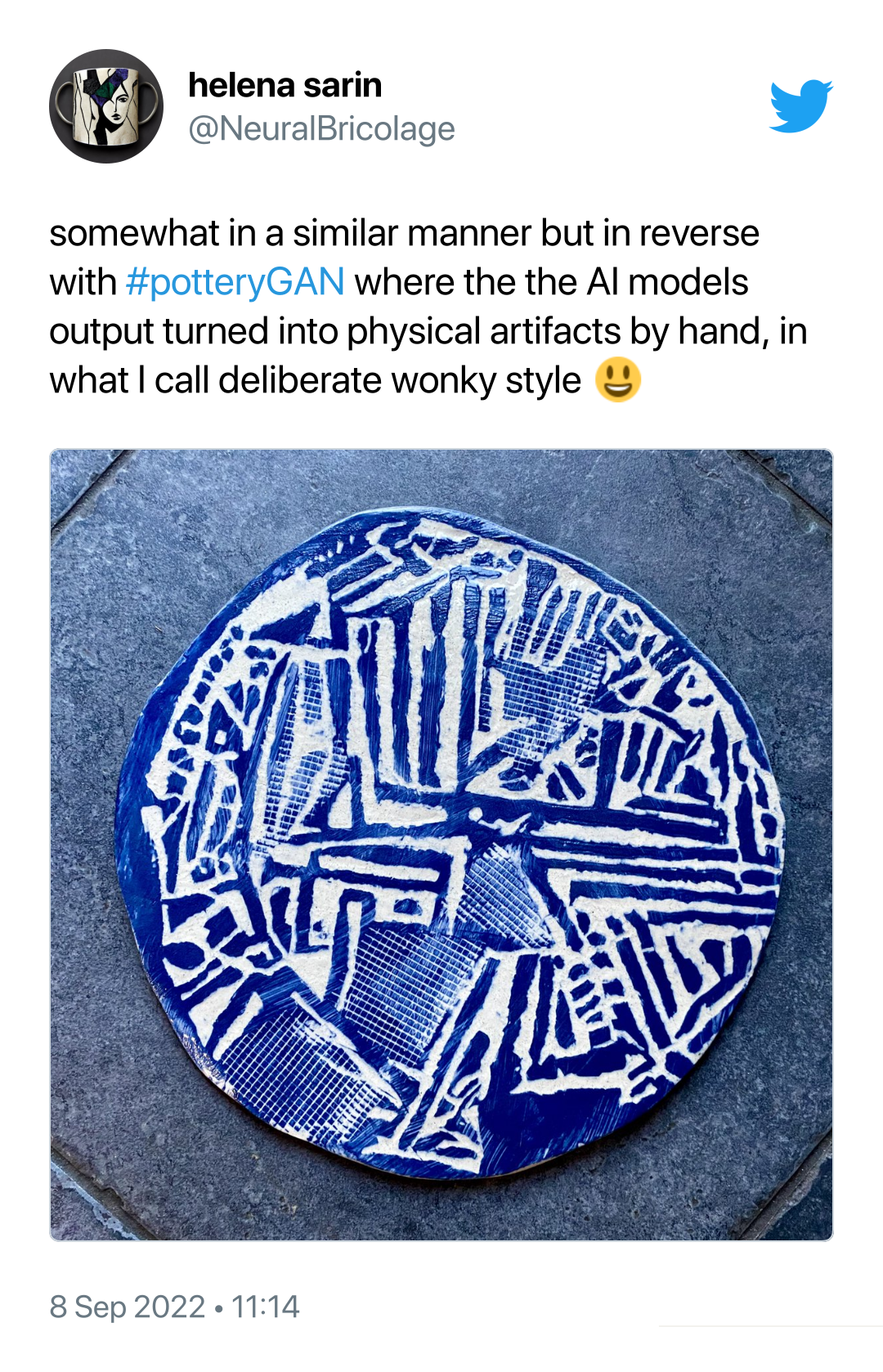}}
        \end{subfigure}
        \caption{Helena Sarin on Twitter}
    \end{figure}

    \item \textbf{Michael Hansmeyer.} He is a postmodern architect who uses GANs to bring fresh perspective to rules-based architecture designs. He has used it to make columns, theaters, etc. The process allows for more artificial serendipity -- the happy accidents and novel ideas that normally take time to stumble upon. His work can be found at \url{https://www.michael-hansmeyer.com/}

    \begin{figure}[H]
    \centering
        \begin{subfigure}{.49\textwidth}
          \centering
          \includegraphics[width=\linewidth]{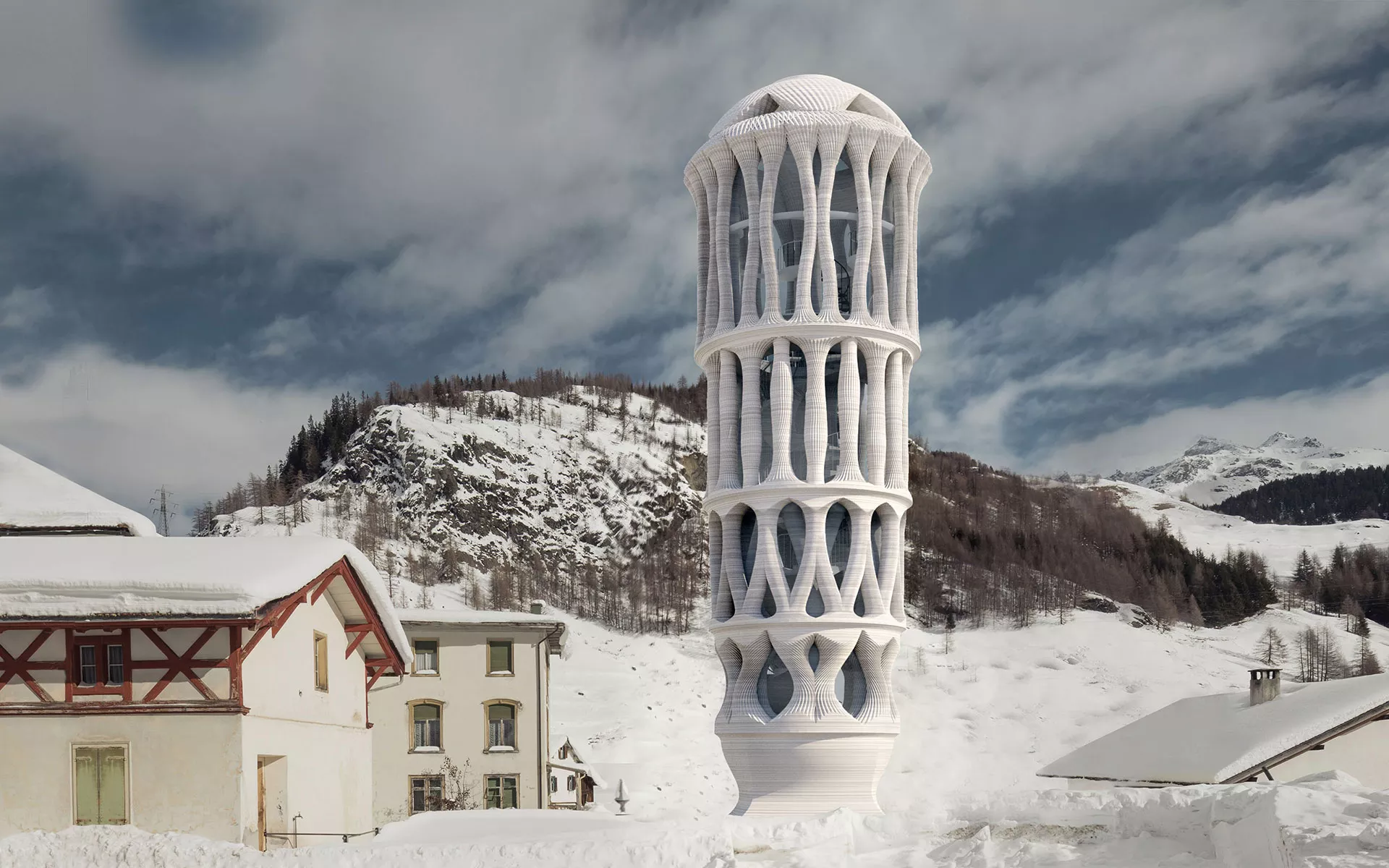}
        \end{subfigure}%
        \hfill 
        \begin{subfigure}{.49\textwidth}
          \centering
          \includegraphics[width=\linewidth]{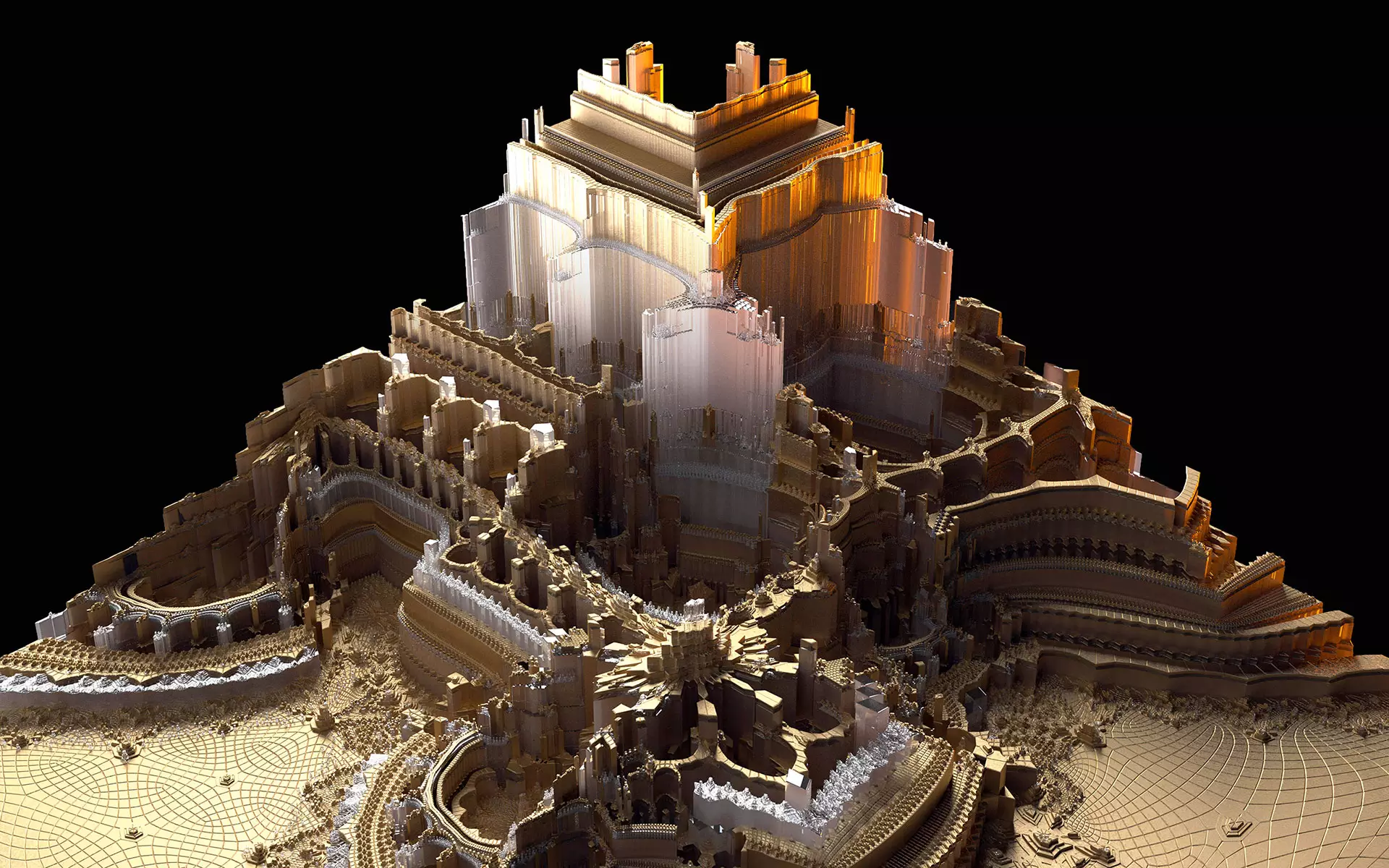}
        \end{subfigure}%
        % \hfill
        % \begin{subfigure}{.33\textwidth}
        %   \centering
        %   \includegraphics[width=\linewidth]{images/michael2.png}
        % \end{subfigure}
        \caption{Architectural work by Michael Hansmeyer}
    \end{figure}

    \item \textbf{Refik Anadol.} He is a Turkish-American new media artist and designer. His projects consist of data-driven machine learning algorithms that create abstract, dream-alike environments. He essentially uses the unrealistic, hallucination-like outputs of GANs to create dreamlike environments and art installations. His work can be found at \url{https://refikanadol.com/}

    \begin{figure}[H]
    \centering
        \begin{subfigure}{.49\textwidth}
          \centering
          \includegraphics[width=\linewidth]{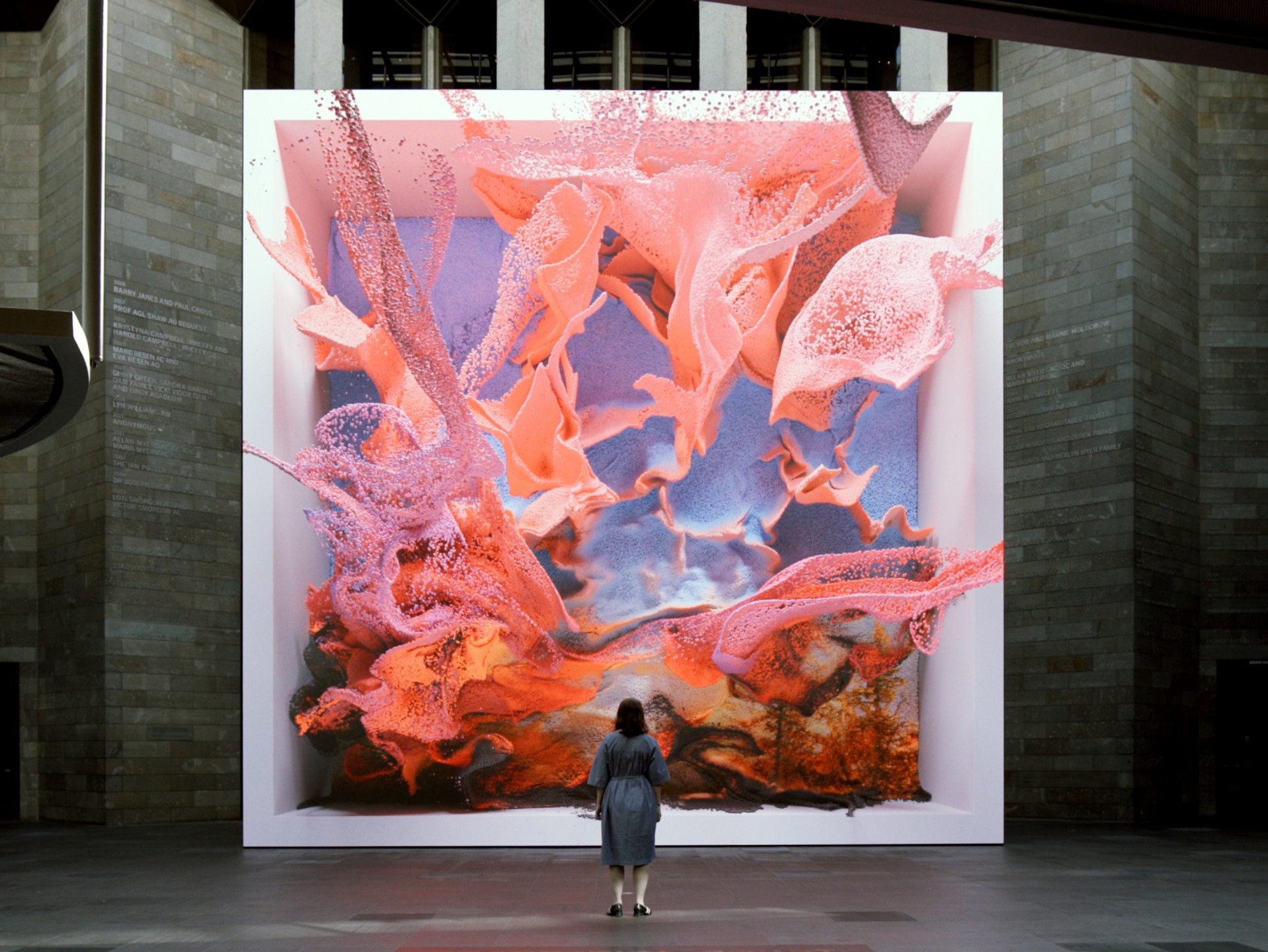}
        \end{subfigure}%
        \hfill 
        \begin{subfigure}{.49\textwidth}
          \centering
          \includegraphics[width=\linewidth]{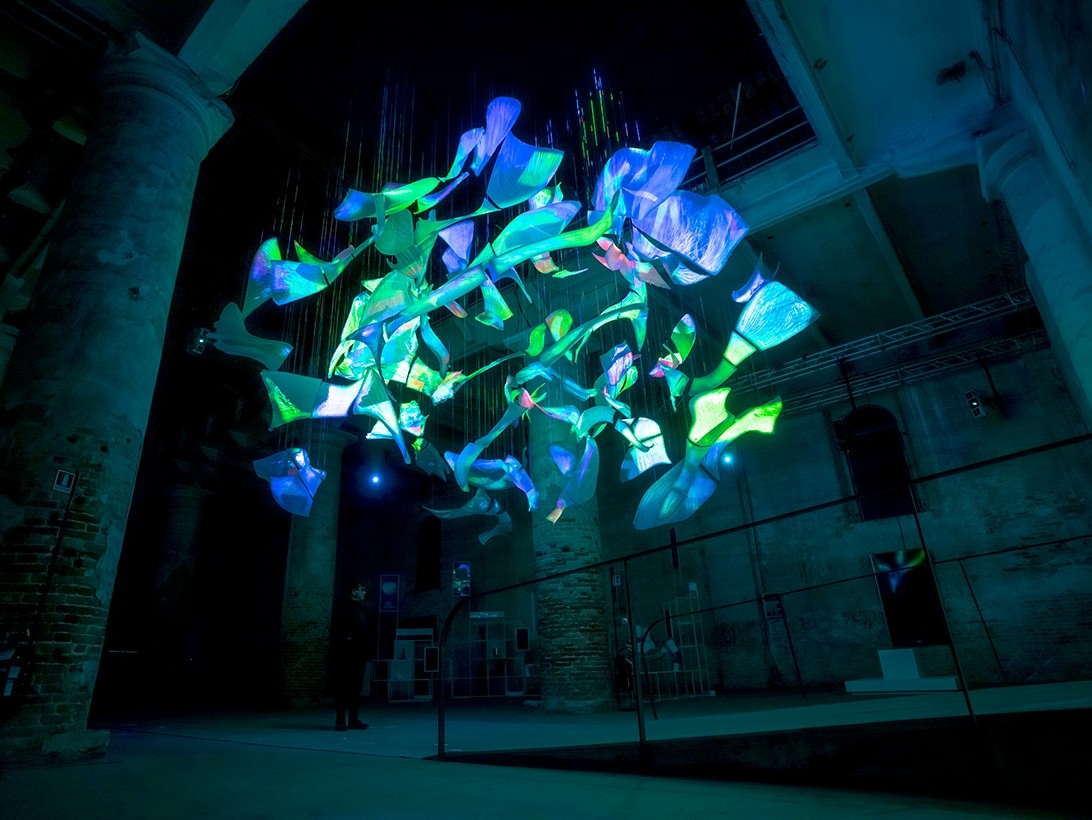}
        \end{subfigure}%
        % \hfill
        % \begin{subfigure}{.33\textwidth}
        %   \centering
        %   \includegraphics[width=\linewidth]{images/michael2.png}
        % \end{subfigure}
        \caption{Dreamscapes by Refik Anadol}
    \end{figure}

\end{itemize}

% \subsection{Modern Instances of AI Replacement of Graphic Designers}

\subsection{HeLa Cells: An analogous debate from Biology}

Henrietta Lacks is known as ``immortal'' for a reason – though she died of cervical cancer at age 30, scientists have used her remarkable cells countless times since. HeLa cells, that never stop dividing and hence are functionally immortal, have played a role in some of the most important medical advancements of our time. They were used to develop the polio vaccine, chemotherapy and cloning technology, among others. However, the original cells that started the immortal HeLa cell line were taken from her without her consent or the awareness of her family. Now her family is demanding compensation from Johns Hopkins University who first took the cells. \footnote{\url{https://www.smithsonianmag.com/smart-news/claims-henrietta-lacks-controversy-far-from-over-180962185/}}

The HeLa cell controversy is yet another cautionary tale about the dangers of cutting out human creators in the pursuit of technology and a lesson in ethics, privacy and consent in technological progress.

\end{document}